\documentclass{sigplanconf}

\usepackage{amsmath}
\usepackage{graphicx}
\usepackage{wrapfig}

\begin{document}

\setlength{\pdfpageheight}{\paperheight}
\setlength{\pdfpagewidth}{\paperwidth}





\titlebanner{banner above paper title}        
\preprintfooter{short description of paper}   

\title{Image Classification Using CNN For Traffic Signs In Pakistan}
\subtitle{}

\authorinfo{Abdul Azeem Sikander}
           {National University of Computer and Emerging Sciences, Islamabad, Pakistan}
           {azeemsikander@hotmail.com}
\authorinfo{Hamza Ali}
           {National University of Computer and Emerging Sciences, Islamabad, Pakistan}
           {i192065@nu.edu.pk}

\maketitle

\begin{abstract}
The autonomous automotive industry is one of the largest and most conventional projects worldwide, with many technology companies effectively designing and orienting their products towards automobile safety and accuracy. These products are performing very well over the roads in developed countries. But can fail in the first minute in an underdeveloped country because there is much difference between a developed country environment and an underdeveloped country environment. The following study proposed to train these Artificial intelligence models in environment space in an underdeveloped country like Pakistan. The proposed approach on image classification uses convolutional neural networks for image classification for the model. For model pre-training German traffic signs data set was selected then fine tuned on Pakistan's dataset. The experimental setup showed the best results and accuracy from the previously conducted experiments. In this work to increase the accuracy, more dataset was collected to increase the size of images in every class in the data set. In the future, a low number of classes are required to be further increased where more images for traffic signs are required to be collected to get more accuracy on the training of the model over traffic signs of Pakistan's most used and popular roads motorway and national highway, whose traffic signs color, size, and shapes are different from common traffic signs.
\end{abstract}




\section{Problem statement}
Develop an artificial intelligence model to predict traffic rules in Pakistan using neural networks by classifying traffic symbols. Identify and classify road signs in different weather conditions, where many road signs are blurry and dark. Panels can appear upside down and make it difficult to understand the patterns. Train model with simple restrictions in Pakistan and other underdeveloped countries to transition to complex constraints. Below is a clear explanation of the problem

\begin{itemize}
\item	Develop a model to predict the traffic rules of Pakistan using convolutional networks by classifying traffic signs images.
\end{itemize}

\section{Introduction}

Image classification is a process of training machine learning models to make them able to classify the data in the images with text \cite{yee2013image}. Many scientists and researchers are already making efforts to get more and more accuracy in this field \cite{lu2007survey}. Recently many experiments were conducted on autonomous robots using image classification. These experiments found several positive results that encouraged researchers to do more research in this area to increase accuracy. After all, autonomous vehicles are the present for the future of the world of robotics and automation \cite{nadeem2018transfer}. Currently, autonomous vehicles are one of the biggest industries in which many well-known industrial products are available in the market. An autonomous vehicle is a driverless vehicle used by a well-trained and perfect artificial intelligence model to drive a car like humans while observing all the rules and regulations on the road and discovering objects on the road to avoid them and determines the route to the destination. The driver-less vehicle is nothing more than a miracle that has existed since cars are economically available until the 1920s. The first step towards self-driving cars was taken in the 1980s by Carnegie Mellon University as part of the NAVLAB project \cite{jochem1995pans}. 

\begin{figure}[h]
    \centering
    \includegraphics[scale=0.25]{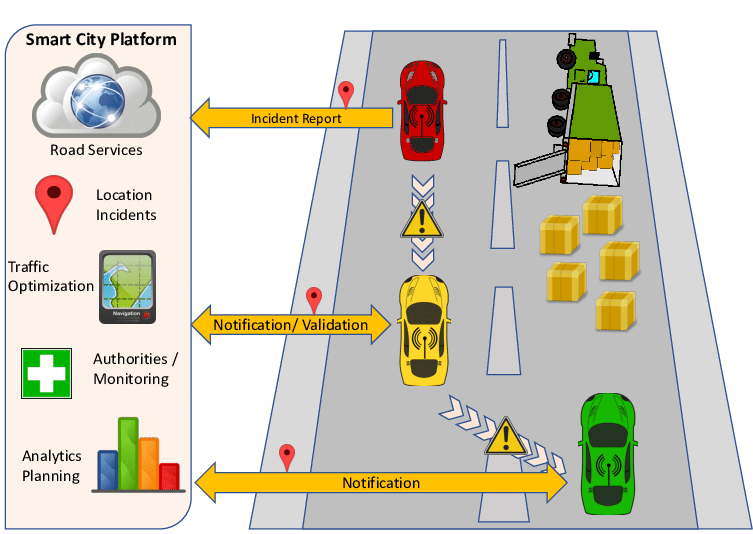}
    \caption{Diagram Explaining scenario of autonomous car activity on the road in case of objects detection and avoidance on the road \cite{Karnouskos}}
\label{fig:mesh1}
\end{figure}
In Fig. 1, the scenario shows the detection of an object by an unmanned car with the truck on the road and the indicated lane closed. Just like in America, these multiple incidents were recorded when self-driving cars collided with trucks on the road.
Later, the world entered a new era of self-driving cars, with several major industrial car companies involved in advancing research on self-driving cars. e.g. For example, among the latest technology companies, Google and the electric car manufacturer Tesla have developed their self-driving car, successfully tested it and offered space to its customers. Where Tesla Model S is very popular in the market which was introduced in June 2012 market but got popular after introducing the semi-auto pilot feature in October 2014 that allows the vehicle to operate semi-autonomously \cite{Lawler}. Self-driving cars include sensors, radars, and GPS systems to detect the environment on the road, obstacles, roads, objects, pedestrians, and other vehicles to ride safely. While detecting the environment one of the most important things which also requires to be detected is traffic sign, rules, warning signs speed limit of the roads, traffic signals and other boards for direction and indication. These things are very important to detect and good accuracy is required to detect these signs from a specific distance to avoid any collision and mistake on the road.

The Tesla Model S is now one of Tesla's most popular automakers, with highly regarded and positive results around the world. But the main focus of the study is to drive these popular self-driving cars in a developed environment \cite{qamarrelationship} and to drive on well-maintained roads where everyone on the road follows the rules well and the violation is sanctioned by the transport authorities, which is good compliance. , but in countries like Pakistan, if these cars are introduced, they will fail at the first corner because the environment and atmosphere of the road in Pakistan are very different from those in well-developed countries.

\begin{figure}[h]
    \centering
    \includegraphics[scale=0.13]{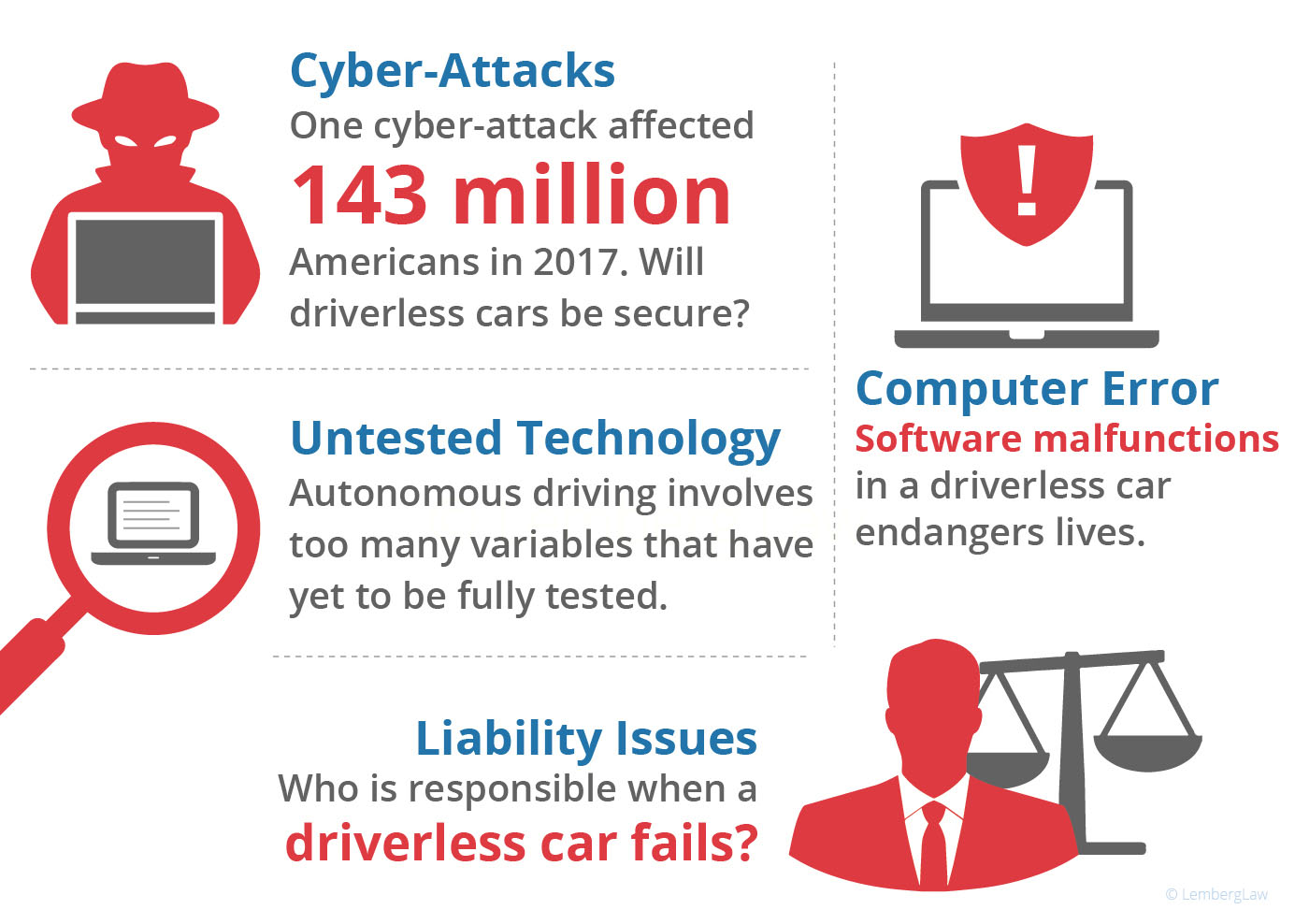}
    \caption{Diagram showing the cyber attack numbers and errors which can occur in the computer systems \cite{NtAVMSF} }
\label{fig:mesh2}
\end{figure}

In this work, an artificial intelligence model has been developed for Pakistani road signs driving cars on Pakistani roads as the computer and autonomy industry grows every day and shows great progress and change. One day in the world, these independent cars and technologies will enter the Pakistani environment. \cite{zafar2018deceptive} These models need to be prepared for the environment of Pakistan and other underdeveloped countries. This is another attempt to help robots and self-driving cars train their models in an undeveloped road environment.  

During this work, Pakistan's traffic sighs data-set was collected. This data set includes all the traffic sign which are common in developed countries and some other which includes signs according to Pakistan's environment. In Pakistan, traffic sign's color, shape, and size are different as compared to the German traffic sign data set, which was selected for the training purpose of the model which was required to train over Pakistan traffic signs. German traffic signs data-set was the pre-training process of the model where the model was fined tuned over the Pakistan traffic signs to get the accuracy over Pakistan's traffic signs dataset. to differ this work from previous work Convolutional neural network was selected for the image classification and the dataset size was increased by adding more traffic signs images into the previously gathered dataset.

\textbf{Motivation:} Autonomous cars are getting very important in the present and the future of robotics in the artificial intelligence field where some models in the future can also be trained for different languages like Urdu is the common language in Pakistan and many traffic signs in Pakistan are also containing only Urdu language as the traffic instruction. So this study can further see text detection in traffic signs. \cite{arshad2019corpus} \cite{majeed2020emotion} \cite{nacem2020subspace} \cite{awan2021top} \cite{naeem2020deep}. In the recent decade, artificial intelligence got a lot of success and provided many useful applications. In developed countries, autonomous cars already ravishing over the roads and showed positive results on roads. But underdeveloped countries are still facing many issues where present AI models will face a lot of difficulties due to lack of traffic rules and proper roads. This work is planning of training AI model over the traffic rules and road situations of Pakistan. Where in Pakistan people do not show a positive response towards traffic rules due to unawareness. The autonomous car industry is developing with great success all around the world. one day this success will cause infiltration of this technology in underdeveloped countries where these countries and the technology both will face these problems. to avoid these problems before we require to train our models and systems according to the current environment or to change the whole structure according to the technology \cite{zafar2020search}. Both will cause a lot of effort and financial issues so before the time will come we should train our models according to the current environment and this is the main motivation for this work.

\begin{figure}[h]
    \centering
    \includegraphics[scale=0.13]{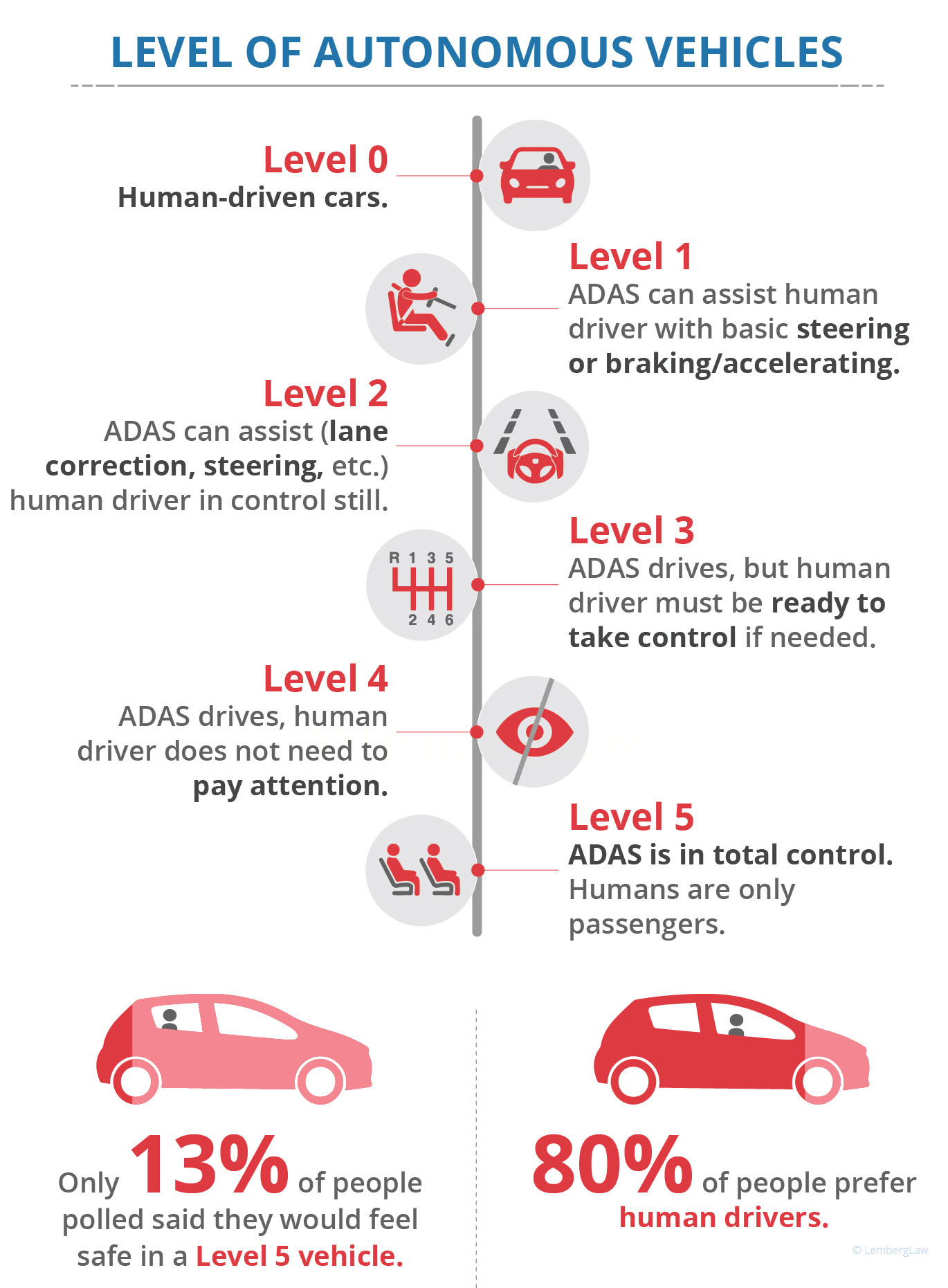}
    \caption{Diagram of different levels in autonomous cars were poll was taken by people and 13 percent feel safe in autonomous cars with out human drivers and other 80 percent feel safe with human drivers. \cite{NtAVMSF} }
\label{fig:mesh3}
\end{figure}

\begin{figure}[h]
    \centering
    \includegraphics[scale=0.13]{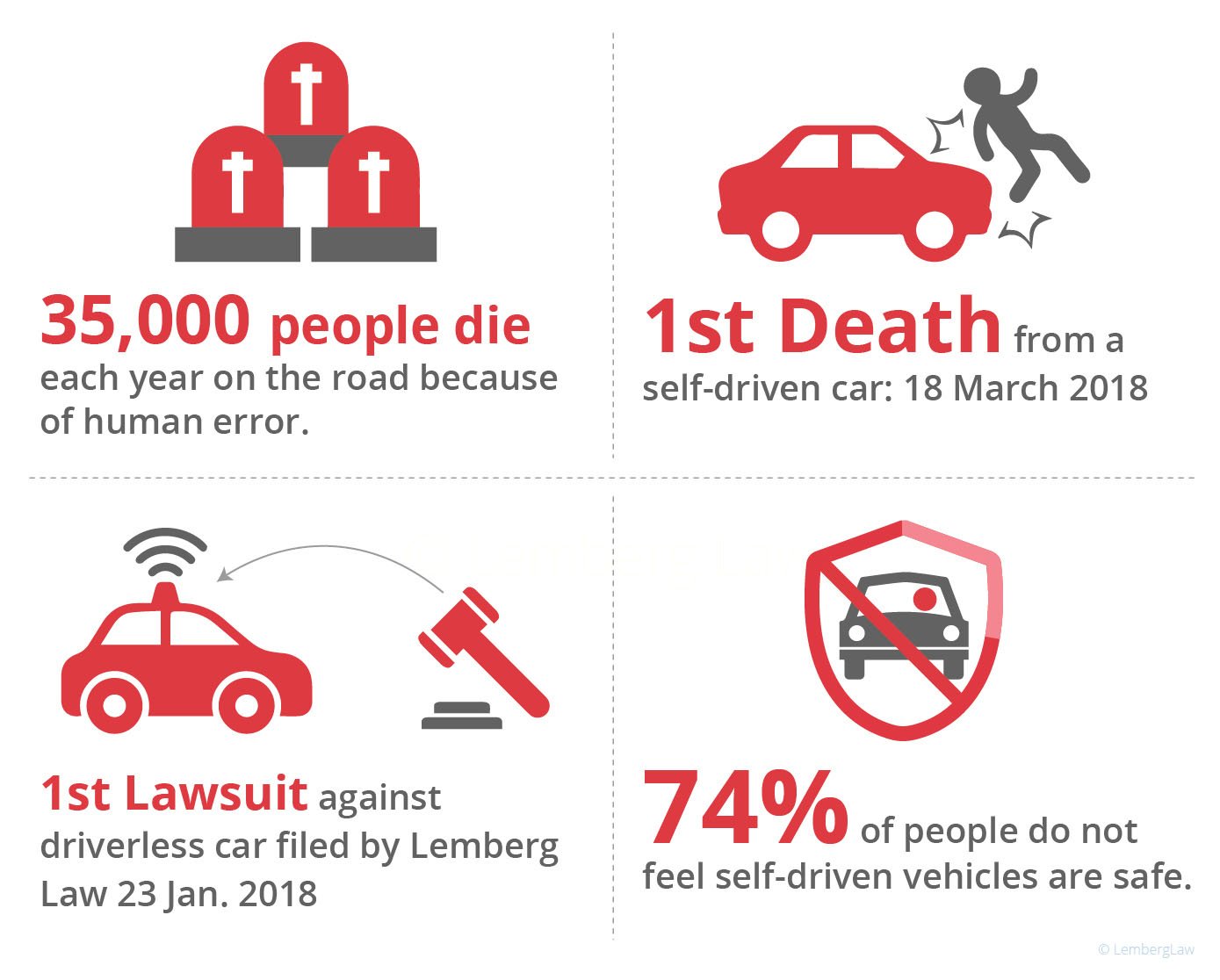}
    \caption{Diagram showing the information of the world wide death rate and first car accident by autonomous car which caused lawsuit on autonomous car and the percentage of people opinion about autonomous cars \cite{NtAVMSF} }
\label{fig:mesh4}
\end{figure}

Currently, Tesla motors based in America is providing the best products in autonomous cars where their cars are also trained for foreign countries' traffic environment. Their models will fail in Pakistan because Pakistan’s traffic is not in a good flow on the roads. On the other hand, where developed countries have strict traffic rules and proper flow still in 2018 Uber self-drive cars hit 49-year-old women in Tempe, Arizona. So launching these cars in Pakistan and other underdeveloped countries is still very dangerous because they will crash in the first second.

As we all know every year 35 thousand people die in road accidents due to their own mistakes and other's mistakes too. The first accident caused by self-driving vehicles was the death of 49-year-old women already mentioned in the above paragraph was the first case in which the first lawsuit was filed against self-driving cars to stop these cars from wandering over the roads. due to that incident, 74 percent of the population of the world thinks that autonomous cars are not a safe option to be implemented on the road \cite{badue2020self}.To get people's trust over self-driving cars researchers need to train their models over multiple things and multiple environments. \newline

\textbf{Background:} Today, artificial intelligence (AI) is present all over the world and continues to grow rapidly. Artificial intelligence works in the following areas: optical character recognition, handwriting recognition, speech recognition, face recognition, artificial creativity, computer vision, virtual reality, and imaging \cite{arora1986uses}. From all the above-mentioned fields, this study lies in image processing where AI is used to detect and recognize objects from images and classify them into different known classes \cite{harchaoui2007image}. There are many well-known algorithms and techniques are available for image classification but Use of neural networks for image classification is one of the best techniques in AI \cite{lu2007survey}. One of the most popular deep learning algorithms in neural networks is the Convolutional Neural Network (CNN). CNN takes the input images and classifies them into various priority classes known to the algorithm. CNN can distinguish between objects found in one image or different images where CNN needs very little training in the dataset \cite{albawi2017understanding}.

\begin{figure}[h]
    \centering
    \includegraphics[scale=0.25]{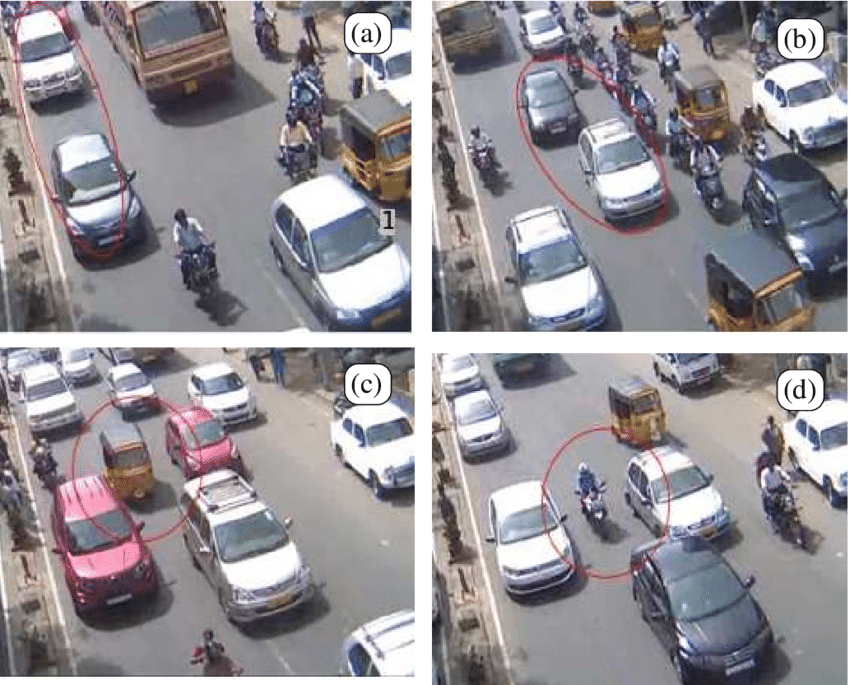}
    \caption{A CCTV footage showing bad behavior of drivers on the road which is the main problem ahead in the field of autonomous cars \cite{Kanagaraj}.}
\label{fig:mesh5}
\end{figure}

Many researchers used CNN in their studies and developed new techniques to increase the accuracy of CNN. Where CNN was also used by experts for traffic signs classification, Which was very useful in autonomous cars. From the start to now many studies were conducted to make autonomous cars more efficient and some popular companies like Tesla successfully developed their product and selling them in the market with high customer satisfaction \cite{farooq2019melta}. These trained AI CNN models are very accurate in their work for detecting and recognizing objects on the road. These models are very well trained on developed countries' roads. In developed countries, many AI successful applications are still working and very accurate in their work. But still, there is a roam to make these models more accurate for the clean and developed country environment. In underdeveloped countries like Pakistan and India, there is still a lot of room available for AI to be implemented even already developed AI models are required to be trained in these underdeveloped countries. Autonomous cars and robotics are one of the hot research areas and many business firms are already developing a lot of AI applications to boost profit. One day for this purpose these firms will take their technology to underdeveloped countries where these successful AI models will fail and their new training will be required according to the new environment. In the previous few years America based motor company Tesla gained a lot of customer attraction and share’s value in the stock market for designing autopilot vehicles and got the promising result in autopilot \cite{bangash2017methodology} . These autonomous cars are designed by implementing physical traffic rules and road sense in the AI model to control the car while driving. The accuracy of those AI models was promising in those countries where traffic rules are very strict and the roads are perfect.  But the biggest challenge is to implement these models in underdeveloped countries where traffic rules are not strict and road conditions are bad people do not like to follow traffic rules properly just like figure 5.

\section{Related work}

According to \cite{wali2015automatic} the first step was taken by Japanese researchers in 1984 for the detection of traffic signs in autonomous cars. Traffic signs detection was also done by image processing techniques \cite{malik2014detection} these state-of-the-art techniques have some limitations in image processing which can be fatal for traffic signs detection. some studies were proposed to done a lot of work on image classification using convolutional neural networks \cite{yim2015image} to get accuracy on the image classification and detect images without any error and exception. The same technique was also used in traffic sign classification \cite{aghdam2016practical}. Convolutional neural networks were first introduced in 1989 with LeNet-1 \cite{lecun1995comparison} and first time used in image processing by lecun et al \cite{lecun1995convolutional} in 1995. Then Convolutional neural networks played a very important role in the field of computer vision and became state-of-the-art for the current techniques. Many studies were conducted for traffic signs of different countries and successfully implemented into autonomous cars and many companies employed these models to use in their autopilot cars. one of the most popular companies is Tesla motors which are currently working on Artificial intelligence with a great pace and successfully selling their products in the market. These models are trained to detect and recognize objects on the road.

 In previous studies, all this work was done on the traffic rules and signs for many countries, but in Pakistan, \cite{nadeem2018transfer} same work was done on signs recognition using Convolutional Neural Networks and transfer learning. In this study, the authors implemented a model using CNN to recognize traffic signs. They trained their model over German Traffic Sign Recognition Benchmark (GTSRB) and then fine-tuned over Pakistan traffic signs and tested on the data set collected from Pakistan. In this study AI model provided very low accuracy on Pakistan's data set, the main reason for this accuracy is a very amount of data set availability because the main challenge for this work was data set collection which was very low from Pakistan. The accuracy of this work can be improved by providing huge data set collection from Pakistan and the model training over signs and rules of traffic. In Pakistan, the traffic environment is very different from other countries which is one of the biggest challenges to face in autonomous car introduction in Pakistan \cite{alvi2017ensights}.

 In another work \cite{wahab2018audio} authors used audio augmentation in which the model was trained on a traffic sign and in testing model recognizes sign and tells the driver about the signs using audio. the accuracy of this work was also very low. according to the authors, the road condition of Pakistan is very bad where vehicles produce a lot of vibration while traveling over the roads. this is also the biggest challenge to train AI models according to Pakistan's road conditions because broken roads and cracks in the roads cause blurriness in the sign recognition. According to the authors, these are not the only challenges in Pakistan as the traffic flow of Pakistan is also not in a manages manner which can also fail auto cars in Pakistan. the previous work in Pakistan is on sign recognition but this study is a proposal for these signs classification model implementation.

\section{Proposed Approach}

In this work, the image classification technique was used to train the AI model on the traffic sign of Pakistan using CNN. The Preprocessing of the model was done over the Germany Traffic Sign Recognition Benchmark (GTSRB),  after preprocessing the trained model was fine-tuned on Pakistan’s collected dataset of traffic signs.

\begin{figure}[h]
    \centering
    \includegraphics[scale=0.19]{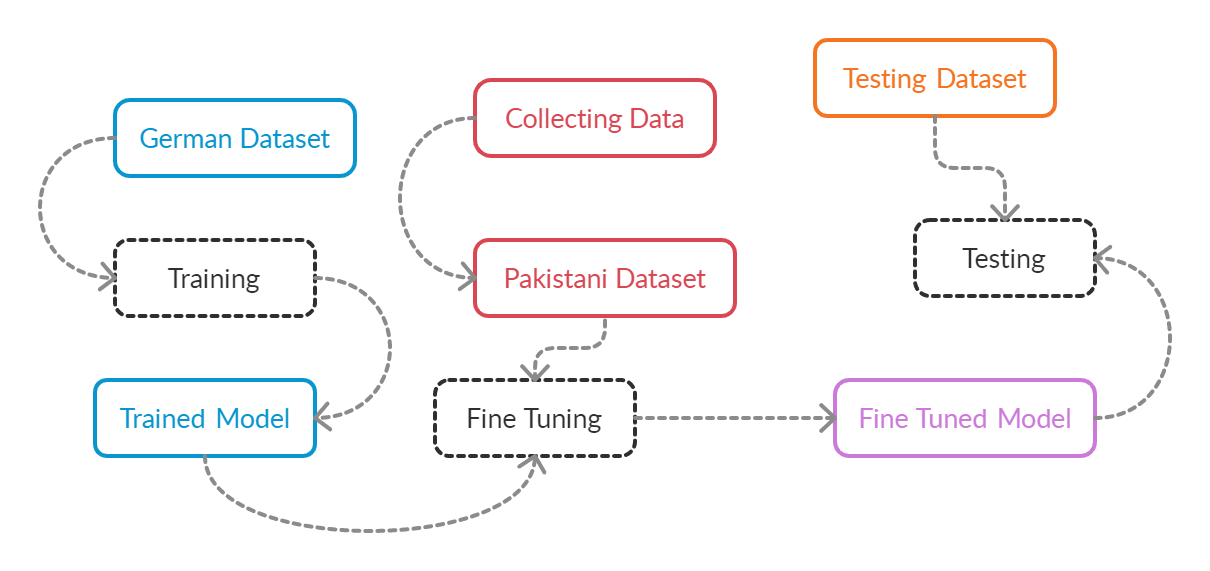}
    \caption{Data flow diagram to show the flow of the proposed approach used in this work to train model for the traffic signs of Pakistan}
\label{fig:mesh6}
\end{figure}

Two steps need to be taken to confirm the study results:
\begin{enumerate}
\item Early training and development of the CNN model and collection of information on German Traffic Signs Recognition Benchmark.
\item Fine-tuning model on Pakistan's traffic signs dataset Using CNN.
\end{enumerate}
\subsection{German Dataset Training and fine adjustment (Phase 1)}
The main phase was carried out using Keras Machine Learning Framework and Tensor Flow.
\subsubsection{Pre-processing:}
The model has been trained in advance to facilitate solidarity. By normalizing the images, we have reduced production times and accuracy, especially when practicing German dataset. Normalization similarly scales the data and balances pixels in the image around a certain point, preventing the slope from evaporating or exploding during the correction process. Another part of the preprocessing was the coding of the names. It represents a clearer understanding of the artificial intelligence model. These are the Keras 'model.fit ()' function and cross entropy categorical loss function \cite{chollet2015keras}.
\subsubsection{CNN Architecture:}
The design model is LeNet Architecture \cite{lecun1998gradient} because it is often used on CNN and is easy to evaluate with good work and fewer images. It requires as little pre-treatment as possible. Image template, figure 7 above included contains, dense layers, maximum polar layers, fallen layers, and fully interconnected layers, which are all convolutional. There are 2 convolutional layers, to minimize images moving beyond organization levels, followed by a maximum grouping level (which is also a control level). Keras machine learning framework was used to develop the model.

\begin{figure}[h]
    \centering
    \includegraphics[scale=0.22]{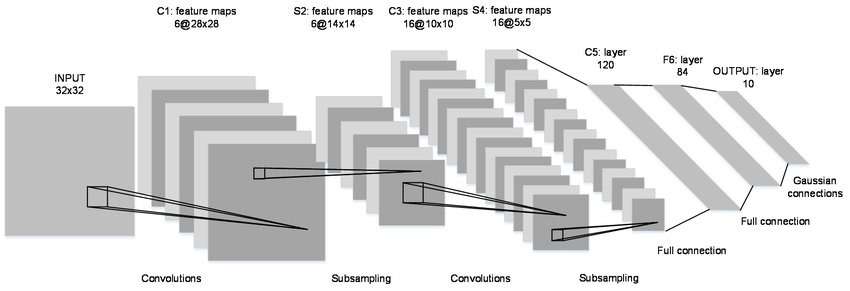}
    \caption{Image of the LeNet-5-Architecture a Convolutional neural network working \cite{TVJKSKJ}.}
\label{fig:mesh7}
\end{figure}
\subsubsection{Training:}
Additionally, training was conducted using several variables 5,20, and 50 were used as appropriate options for comparison The "Categorical Cross-entropy" responses were used as a loss while using "Adam Optimizer \cite{kingma2014adam}" Correction. The ‘model. fit’ function was used returned to many metrics which was very useful for accuracy and loss value for dataset test and validation used later in functions optimized results and processes mentioned previously.

\subsubsection{Regularization:}
Excessive adaptation of information threatening neural networks and their proliferation cycles is a serious problem. To reduce false positive effects, regulatory methods are used to maintain a strategic distance from additional information. Results obtained afteer learning of model on incorporate “unacceptable results” because it is more appropriate and begins where results are produced with less accuracy and great precision. The work strategy was to interrupt, This regulatory process focuses on the idea of "learning less to learn better". Giving up means learning about the burdens and tendencies that prevent mixing. It is known to work better and is basically a modern school standard for deep learning \cite{srivastava2014dropout}.

\subsection{Pakistan Traffic Signs Dataset}
In the second phase of the transfer learning research \cite{asad2020deepdetect}, the standard model was modified from the first phase to capture the key points of the Pakistani dataset. This section also describes how to change the information selection and template design procedures to create a template that best meets our needs.
\subsubsection{Data Collection}

\begin{figure}[h]
    \centering
    \includegraphics[scale=0.5]{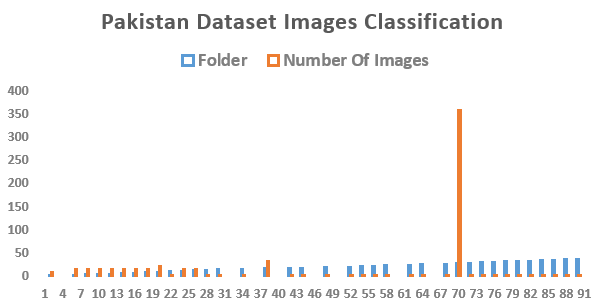}
    \caption{Data chart to show the number of images in pakistan dataset used in this work to train model for the traffic signs of Pakistan}
\label{fig:mesh8}
\end{figure}

Over time, data has been collected and downloaded from different sources like google and Islamabad traffic police website and some cities \cite{dilawar2018understanding} as this data also includes the previous collection which was done by Nadeem et al \cite{nadeem2018transfer} in their work, which includes cities like Quetta, Islamabad and Karachi, Pakistan, but not limited to only these cities as motorway and Pakistan national highway dataset can also be collected. Finally, although the number of images collected as part of the Pakistani material is small, the images described in table 1. are transferred disproportionaly between classes. As shown in Figure 8, It is a Pakistani dataset just like the German dataset. \cite{Houben-IJCNN-2013}

\begin{table}[h!]
\begin{center}

\begin{tabular}{|c|c|c|}
\hline
\textbf{\#} & \textbf{Sign Name}                                                                     & \textbf{Number Of Images} \\ \hline
\textbf{1}  & \textbf{\begin{tabular}[c]{@{}c@{}}Audible warning \\ devices prohibited\end{tabular}} & \textbf{8}                \\ \hline
\textbf{2}  & \textbf{Cattle crossing}                                                               & \textbf{15}               \\ \hline
\textbf{3}  & \textbf{Children crossing}                                                             & \textbf{15}               \\ \hline
\textbf{4}  & \textbf{Dangerous descent}                                                             & \textbf{15}               \\ \hline
\textbf{5}  & \textbf{End 0f 30 Kmh zone}                                                            & \textbf{15}               \\ \hline
\textbf{6}  & \textbf{Falling rock}                                                                  & \textbf{15}               \\ \hline
\textbf{7}  & \textbf{Give Way}                                                                      & \textbf{15}               \\ \hline
\textbf{8}  & \textbf{Go straight ahead}                                                             & \textbf{15}               \\ \hline
\textbf{9}  & \textbf{Go straight or left}                                                           & \textbf{22}               \\ \hline
\textbf{10} & \textbf{Left bend}                                                                     & \textbf{1}                \\ \hline
\textbf{11} & \textbf{Left turn ahead}                                                               & \textbf{15}               \\ \hline
\textbf{12} & \textbf{Light signals}                                                                 & \textbf{15}               \\ \hline
\textbf{13} & \textbf{Major cross road}                                                              & \textbf{1}                \\ \hline
\textbf{14} & \textbf{\begin{tabular}[c]{@{}c@{}}Maximum speed \\ limit 5 Kmph\end{tabular}}         & \textbf{1}                \\ \hline
\textbf{15} & \textbf{\begin{tabular}[c]{@{}c@{}}Minor cross \\ road from left\end{tabular}}         & \textbf{1}                \\ \hline
\textbf{16} & \textbf{\begin{tabular}[c]{@{}c@{}}Motorway Police \\ Dataset\end{tabular}}            & \textbf{33}               \\ \hline
\textbf{17} & \textbf{No entry}                                                                      & \textbf{1}                \\ \hline
\textbf{18} & \textbf{\begin{tabular}[c]{@{}c@{}}No entry for \\ goods vehicles\end{tabular}}        & \textbf{1}                \\ \hline
\textbf{19} & \textbf{\begin{tabular}[c]{@{}c@{}}No Entry for \\ motor vehicle\end{tabular}}         & \textbf{1}                \\ \hline
\textbf{20} & \textbf{No left turn}                                                                  & \textbf{1}                \\ \hline
\textbf{21} & \textbf{No parking}                                                                    & \textbf{1}                \\ \hline
\textbf{22} & \textbf{No right turn}                                                                 & \textbf{1}                \\ \hline
\textbf{23} & \textbf{\begin{tabular}[c]{@{}c@{}}No stopping \\ (Clearway)\end{tabular}}             & \textbf{1}                \\ \hline
\textbf{24} & \textbf{No U-turn}                                                                     & \textbf{1}                \\ \hline
\textbf{25} & \textbf{\begin{tabular}[c]{@{}c@{}}Overtaking \\ prohibited\end{tabular}}              & \textbf{1}                \\ \hline
\textbf{26} & \textbf{Pedestrian crossing}                                                           & \textbf{1}                \\ \hline
\textbf{27} & \textbf{Real Data}                                                                     & \textbf{357}              \\ \hline
\textbf{28} & \textbf{Right bend}                                                                    & \textbf{1}                \\ \hline
\textbf{29} & \textbf{Road works}                                                                    & \textbf{1}                \\ \hline
\textbf{30} & \textbf{Slippery roads}                                                                & \textbf{1}                \\ \hline
\textbf{31} & \textbf{Slow}                                                                          & \textbf{1}                \\ \hline
\textbf{32} & \textbf{Steep ascent}                                                                  & \textbf{1}                \\ \hline
\textbf{33} & \textbf{Stop}                                                                          & \textbf{1}                \\ \hline
\textbf{34} & \textbf{Turn to the left}                                                              & \textbf{1}                \\ \hline
\textbf{35} & \textbf{Two way traffic}                                                               & \textbf{1}                \\ \hline
\textbf{36} & \textbf{Un-even road}                                                                  & \textbf{1}                \\ \hline
\textbf{37} & \textbf{U-turn}                                                                        & \textbf{1}                \\ \hline
\end{tabular}
\end{center}
\caption{\label{tab:table-name}Showing data collection of pakistani dataset.}
\end{table}

The 359 images were collected by Nadeem et al \cite{nadeem2018transfer} and 579 total images were gathered around by collected 220 new images from different sources and then physically created to filter out background information from unwanted images. All images are organized according to correspondence that includes the names of the images. With Python code, 579 images were named alphabetically by group name (thus "Bridge Ahead" images are 0-12, while "Zigzag Road Ahead" images are 346-359). These image parts are not clipped same , CNN recognizes only one size image no matter what. Therefore, all images were converted to 32x32 format. These pre-attached images are flattened, stacked on the screen, and tagged with their tutorials so that they are ready to be presented in the preparation module. The general statistical measures remained the same as the German dataset.

\section{Evaluation and Experiments}

The most focused process in the experimentation was the collection of data which plays a very vital role in the implementation and training of AI models. After collecting data from different sources and already collected datasets from previous studies the study was forwarded towards the selection of the benchmarks which also plays a very crucial role in the training of models as these benchmarks were used for the preprocessing of the model. After the selection of benchmark, the model was implemented on the benchmark dataset DTSRB which trained the model on the traffic signs then the fine-tuning of the dataset collected from Pakistan was used to fine-tune the model to test the main accuracy of the model on the Pakistans dataset.
Hardware specifications of the machine are described below which was used in this work to perform whole experiment training and testing.

\begin{table}[h!]
\begin{center}
\begin{tabular}{||c|c||}
\hline
\multicolumn{2}{||c||}{\textbf{Hardware Specifications}} \\ 
 \hline
\textbf{Processor}     & Core i7 6700K 4 GHZ           \\ \hline
\textbf{Ram}           & 8 GB DDR4 2400 MHZ                 \\ \hline
\textbf{Graphic Card}  & Geforce GTX Nvidia 1070 8 GB  \\ \hline
\textbf{Hard Disk}      & 128 GB SSD                    \\ \hline
\textbf{Power Supply}  & Thermaltake 700 Watt          \\ \hline
\end{tabular}

\end{center}
\caption{\label{tab:table-name1}Hardware Specifications of machine used for training Artificial Intelligence model over Traffic signs of Germany dataset and Pakistan's dataset}

\end{table}

For this study, GTRSB is being selected as the benchmark of the model training. "Benchmark for the recognition of Germany road signs" is a multidimensional administrative challenge organized by the IJCNN in 2011. Regular approval of road signs is required to remove the driver support frame and promote a tougher approach to driving. 'Computer, like the identity problem. More than 50,000 images of traffic signals are collected from remote and accurate data sets. It reflects solid forms of symbolic intuition due to distance, light, climatic conditions, and obstacles along the way. Image production involves a series that is considering calculating artificial intelligence without basic knowledge.

\begin{table}[h!]
\begin{center}

\begin{tabular}{||c|c||}
\hline
\multicolumn{2}{||c||}{\textbf{Software Specifications}} \\ 
 \hline
\textbf{Windows} & Windows 10 Professional x64 Bit           \\ \hline
\textbf{Python} & Python 3.8.5                  \\ \hline
\textbf{Juypter Notebook}  & 6.1.4  \\ \hline
\textbf{Conda} & Conda 4.9.2                    \\ \hline
\end{tabular}
\end{center}

\caption{\label{tab:table-name2}Software Specifications of machine used for training Artificial Intelligence model over Traffic signs of Germany dataset and Pakistan's dataset.}

\end{table}

The data set consists of 43 categories, on the other hand, the clock is ticking. Members want to complete two sets of tests with more than 12,500 images. The first results of this series are used in the critical diagnostic phase of the two-sync challenge. The strategy used by high-performing members is in direct conflict with the identification and implementation of image and benchmarks for human transport \cite{stallkamp2011german}\newline
 Specifications of the software are described below which is used for the implementation, training, and testing of the model over the GTRSB and Pakistan's traffic signs data set.
After the selection of datasets and machines, the python code was written to get Pakistan's dataset in shape where GTRSB was already in the shape and not needs any preprocessing. After the implementation of the model on the GTRSB the summary of the CNN was collected over the dataset which is shown below in the figure. In that figure, the CNN layers and the number of parameters per layer are shown.

\begin{figure}[h]
    \centering
    \includegraphics[scale=0.6]{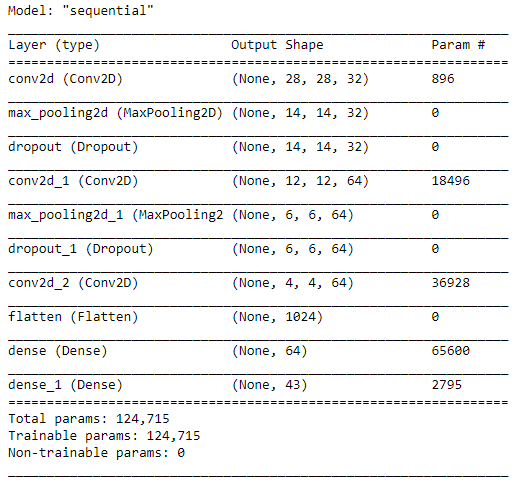}
    \caption{Summary of the convolutional neural network trained model for German Dataset}
\label{fig:mesh11}
\end{figure}

After getting a summary the dataset of GTRSB was divided into a test set and train set where the training and the testing of the model were performed on these split sets. After testing the model showed 98 \% of the test accuracy over the GTRSB which was the preprocessing of the model. The model showed promising accuracy on GTRSB which ables the model to be tested over Pakistan’s dataset.
\begin{figure}[h]
    \centering
    \includegraphics[scale=0.4]{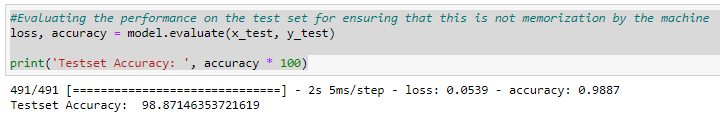}
    \caption{Test Accuracy rate of German Dataset Trained Model}
\label{fig:mesh12}
\end{figure}

As the model showed great accuracy on the GTRSB the graphical representation of the test accuracy is plotted into the form of a graph. The plotted graph of training and validation of the model on GTRSB shows two different lines different in color. The blue line indicates the accuracy of the training and the orange line the accuracy. The training accuracy is more than validation accuracy. The previous numeric number showed the accuracy in one number but the graph shows both accuracies on the same platform which also helps to make a comparison between these accuracies. On the graph, the x-axis is showing the accuracy of the graph and the y axis is showing several epochs which are the number of iterations performed during the training and the validation of the model. In graph accuracy of the model on both training and validation is shown on all the numbers of epochs where from the start before 5 epochs the accuracy of training was below 60\% but the accuracy of validation was above 60. And after 10 epochs the training accuracy crossed the validation accuracy and with some increase or decrease in both accuracies after 15 epochs the training accuracy increased while the validation accuracy started decreasing. The training accuracy finished above 80\% and validation accuracy decreased to below 70\% after 30 epochs which means the model trained itself well on the german data set instead of testing.

\begin{figure}[h]
    \centering
    \includegraphics[scale=0.7]{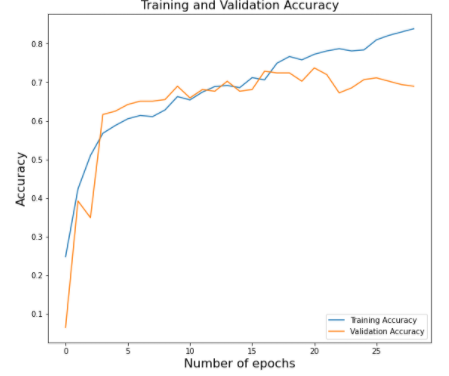}
    \caption{Graph showing the training and validation Accuracy rate on the test data for German Dataset}
\label{fig:mesh13}
\end{figure}

After getting accuracies over the GTRSB model training and validation the loss for both was also calculated which showed some different results over the loss for both sets. The x-axis in the graph shows the loss of both sets and the y axis shows the number of epochs. While in this run before the first epoch the loss of the training set was very high than the loss of the validation set but after the first iteration before the second the training loss minimized below 3 and even lesser than the validation loss. With the increase in the number of iterations, the loss for both sets started decreasing and the training loss gradually decreased as compared to the validation loss where the validation loss slightly started increasing on the twentieth epoch where the validation loss remained on 2 and the training loss decreased below 1 which shows the more accuracy of the model in training set than the validation set.

The above figure no 13. Shows the labels of the actual dataset image and the label of the predicted dataset image. All traffic signs in the image have some issues in the clarity of the sigh where humans can predict the sigh with some ease but in some cases, even humans can find it hard to predict the sign correctly from some distance in bad weather. In the image, some sign is very darker which are very hard to predict but the model predicted signs correct except one sign which is very bad and very hard to tell. This collective image of all signs is just to show the accuracy rate of the model on all kinds of traffic signs images. This prediction image is just for the GTRSB dataset.

\subsection{Pakistan Dataset Results}
Experimental setup provided results to show that neural networks and especially clean-haired convolutional neural networks require a lot of information, otherwise they can be used to create expectations \cite{lecun2010convolutional}. The GTRSB dataset used for the training process of the model is very huge in amount of image count (ca 39000 \cite{lafuente2010decision}). So far, a better description of the model is not enough, and we still need to use the data to process the data expansion process. Similarly, it can be assumed that the Pakistani dataset is very small to benefit CNN training. Then learning exchange was used, which allows the model to solve the problem without a huge number of data. As mentioned above, a model trained with German signs related to the traffic was applied to the Pakistani dataset.

\begin{figure}[h]
    \centering
    \includegraphics[scale=0.6]{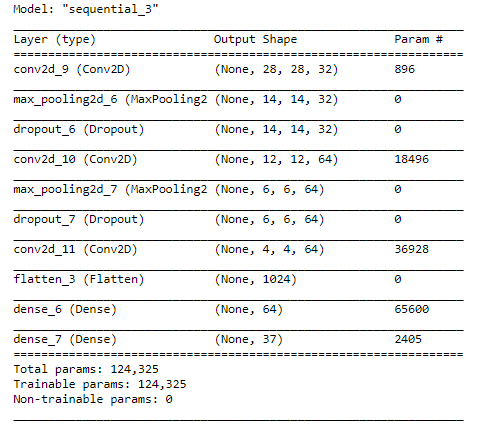}
    \caption{Flow chart showing the Training and Validation Loss rate on the test data for German Dataset}
\label{fig:mesh14}
\end{figure}

\begin{figure}[h]
    \centering
    \includegraphics[scale=0.47]{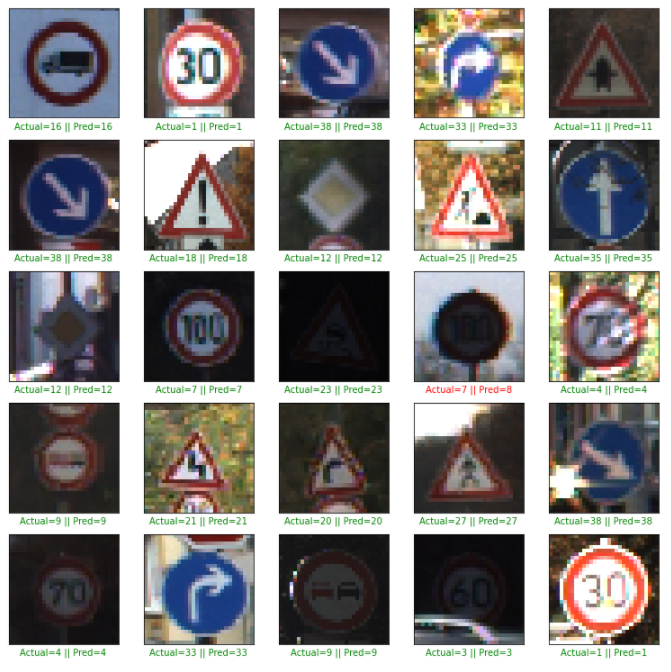}
    \caption{Images Chart showing actual vs predicted labels by the trained model on German Dataset, Green labels showing correct prediction while red labels showing incorrect predictions}
\label{fig:mesh15}
\end{figure}

The following diagram is showing the CNN layers implied on the Pakistan traffic signs dataset and the number of parameters in each layer. Where in the first layer CNN have 896 params but below layers have zero params but the other CNN layer have 18496 layers according to this diagram every Conv2D layer have parameters as compared to other layers but both dense layers have their number s of params which collectively shows the 124325 as the number of total params and the same number as the Trainable Params where all params will be included in the training of the model. This is the main purpose of this graph to just show the total numbers and trainable numbers of parameters in the dataset.

\begin{figure}[h]
    \centering
    \includegraphics[scale=0.7]{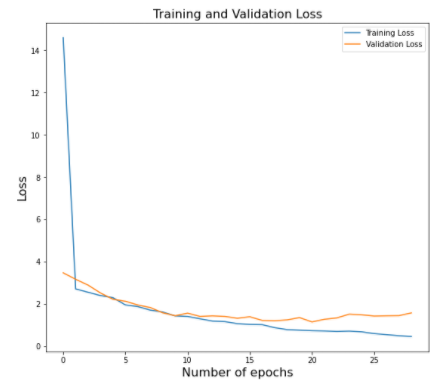}
    \caption{Flow chart showing the training and validation Loss rate on the test data for German Dataset}
\label{fig:mesh16}
\end{figure}

The results of the fine-tuning, was invincible after an increase in the data set the number of images in the data was bound to 579 images. Where more than 50\% of the dataset was containing the collected images from the previously mentioned studies of Nadeem et al. in their experiment due to the low rate of the dataset their accuracy of the model was 41\% which is very low as compared to the other models training accuracy but the main flaw in their research was the amount of dataset. To decrease this flaw more dataset images were added to their data set to perform fine-tuning. After implementing the same CNN models using the same procedure the new model showed a great increase in accuracy, as compared to their model score it increased from 41\% to 83\% which is a promising result by the model. As Nadeem et al mentioned in their study the dataset size was the main factor for the low accuracy in their work. Just by increasing the dataset size, the model increased the accuracy over the same dataset. In this work, the main contribution toward the CNN, image classification, and Autonomous industry of Pakistan is the availability of a good amount of traffic signs dataset and the well-trained model over the available dataset.

\begin{figure}[h]
    \centering
    \includegraphics[scale=0.5]{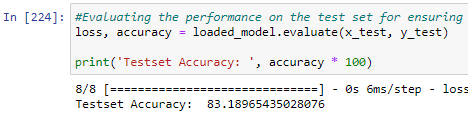}
    \caption{Images Chart showing the actual vs predicated labels by the trained model on German Dataset, Green labels showing correct prediction while red labels showing incorrect predictions}
\label{fig:mesh17}
\end{figure}

The accuracy of the fine-tuned model showed before is simply from the python code representation where the accuracy is showed in the form of a number over the Pakistans traffic sign dataset. While in the below graph the accuracy of this model was shown with some detailed representation in the form of a graph. Where the x-axis in the graph is representing the accuracy of the model over the dataset and the y axis is showing the number of epochs which are the number of iteration of the model on the dataset. The blue line in the graph is representing training accuracy and the orange line is representing test accuracy. From the start, the training accuracy of the model for the Pakistani dataset was low as compared to the test accuracy. Where test accuracy started from 80\% and the training accuracy started from 73\%. With the increase in the number of iterations, the training accuracy started increasing with a great pace as it crossed 90\% but then have some decrease and increase in it and stopped at the 90\% after the finish in the epochs. On the other hand where test accuracy was above training accuracy decreased below training and remains low and decreasing with the number of iterations. Over Pakistan’s dataset, the model training was good as compared to the testing accuracy where the overall accuracy of the model was very good. 

\begin{figure}[h]
    \centering
    \includegraphics[scale=0.7]{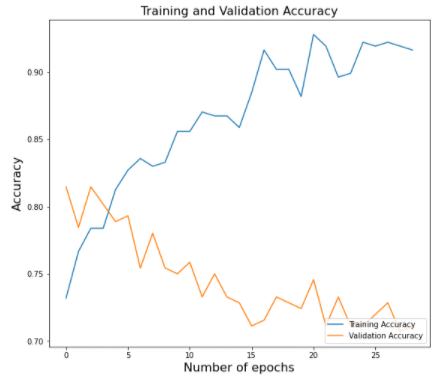}
    \caption{Flow chart showing the Training and Validation Accuracy rate on the test data for Pakistan Traffic Sign Dataset}
\label{fig:mesh18}
\end{figure}

Similar to the previous diagram, which shows the accuracy of the test and practice, this diagram illustrates the loss of the training set and the test set. In the diagram, the x-axis is the loss, while the y-axis is the number of times. Compared to the accuracy table where the training accuracy was higher than the test accuracy, the same is true here if the training set is very small than the loss of the test set.

\section{Conclusion and Future Work}
In this present era, artificial intelligence is overtaking every field of computer science and infiltrated into the lives of every single human being on planet earth.  Now every industry is using computer equipment, and AI is progressing at a great speed. With this kind of progress in artificial intelligence, many biggest industries are now adopting it. One of the biggest industries with AI is robotics and autonomous vehicles where many big names from the industrial line adopted this field and created a great change in the field of AI. With autonomous cars, big running firm Tesla Motors is making new toys and developing a new thing every day. Computer science and AI became very powerful these days and gaining control over every country. At this point, Autonomous cars are trained for developed countries but one day these models will be required to be trained in underdeveloped countries. The purpose of the work was to create a CNN model for recognizing traffic signs in Pakistan. CNN requires large measurements of information for beneficial accuracy.  The CNN model trained for the German data set was refined with the Pakistani data set, the final accuracy of the Pakistani traffic signs was 83\%, which was achieved after 30 iterations, as shown in Figure 16.

the importing thing on pre-training model of over 39,000 images was developed for the dataset produced in Germany, although the calibration model only contains 579 images and is as efficient as what this article focuses on. But still, these images were increased from 359 to this size, where this increase in the size helped to increase the accuracy of the model training instead of testing. The number of images is very important and we can see the difference in the accuracy of the final productions. The general purpose of this study is to provide information on how the lack of images removed during preparation can be highlighted in a similar way to those mentioned in this article.

These results can be considered a failure because we could not collect enough images during the review due to time and money, we only collected 220 images and gathered previously collected data and combined whole data to increase the size but still the size of data was not enough. The most resourceful part is the selection of information for a test based on machine learning. Due to the inaccessibility of the information, it was not possible to create a model that could summarize too much accurately. However, the accuracy of the relevant specimen for the set data size was 83\%.

If you work with a small dataset and provide such accuracy, you are ready to receive traffic lights, like in Pakistan and other unfamiliar/agricultural countries where these cards are available, but this integrated hardware is not available. Understandable data transfer from other datasets, which facilitates computation and financing, maintains various AI combination computations to identify and maintain traffic signs and further processing damage.

The results suggest that the results can be significantly improved with preparation/approval accuracy. These improvements correspond directly to the number of images in the data set, and the extension of this threshold is preferably the result of greater accuracy. The study also identified 35- classes. It develops later than usual because collecting more images creates more road signs.

\begin{figure}[h]
    \centering
    \includegraphics[scale=0.7]{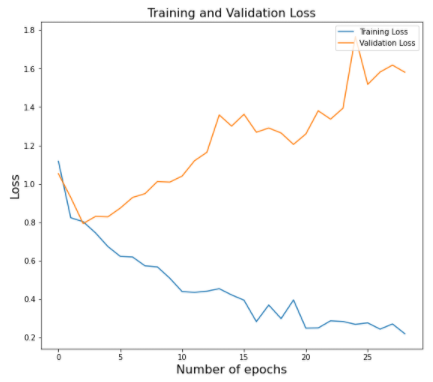}
    \caption{Flow chart showing the Training and Validation Loss rate on the test data for Pakistan Traffic Sign Dataset}
\label{fig:mesh19}
\end{figure}



\bibliographystyle{abbrvnat}
\bibliography{cnn}

\end{document}